\documentclass[hidelinks]{article}

\usepackage{arxiv}

\usepackage{doi}

\usepackage[utf8]{inputenc} 
\usepackage[T1]{fontenc}    
\usepackage{hyperref}       
\usepackage{url}            
\usepackage{booktabs}       
\usepackage{amsfonts}       
\usepackage{nicefrac}       
\usepackage{microtype}      
\usepackage{cleveref}       
\usepackage{graphicx}
\usepackage{natbib}
\usepackage{doi}

\title{A Seven-Layer Model for Standardising AI Fairness Assessment}

\author{ \href{https://orcid.org/0000-0003-4553-5861}{\includegraphics[scale=0.06]{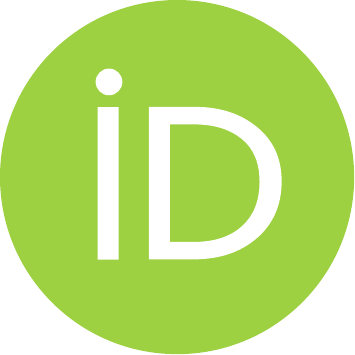}\hspace{1mm}Avinash ~Agarwal} \\
  Telecommunication Engineering Centre\\
  Ministry of Communications\\
  Government of India\\
  New Delhi, India \\
  \texttt{avinash.70@gov.in} \\
  \And
  \href{https://orcid.org/0000-0003-3193-1521}{\includegraphics[scale=0.06]{orcid.pdf}\hspace{1mm}Harsh ~Agarwal} \\
  Adobe Inc.\\
  Noida, India \\
  \texttt{harshaga@adobe.com} \\
}

\renewcommand{\undertitle}{}
\renewcommand{\headeright}{}
\renewcommand{\shorttitle}{A Seven-Layer Model for Standardising AI Fairness Assessment}

\hypersetup{
pdftitle={A Seven-Layer Model for Standardising AI Fairness Assessment},
pdfsubject={q-cs.AI},
pdfauthor={Avinash ~Agarwal, Harsh ~Agarwal},
pdfkeywords={Artificial intelligence, audit, ethics, fairness, standardisation, trustworthy AI},
}

\begin{document}
\maketitle

\begin{abstract}
Problem statement: Standardisation of AI fairness rules and benchmarks is challenging because AI fairness and other ethical requirements depend on multiple factors such as context, use case, type of the AI system, and so on. In this paper, we elaborate that the AI system is prone to biases at every stage of its lifecycle, from inception to its usage, and that all stages require due attention for mitigating AI bias. We need a standardised approach to handle AI fairness at every stage.

Gap analysis: While AI fairness is a hot research topic, a holistic strategy for AI fairness is generally missing. Most researchers focus only on a few facets of AI model-building. Peer review shows excessive focus on biases in the datasets, fairness metrics, and algorithmic bias. In the process, other aspects affecting AI fairness get ignored.

The solution proposed: We propose a comprehensive approach in the form of a novel seven-layer model, inspired by the Open System Interconnection (OSI) model, to standardise AI fairness handling. Despite the differences in the various aspects, most AI systems have similar model-building stages. The proposed model splits the AI system lifecycle into seven abstraction layers, each corresponding to a well-defined AI model-building or usage stage. We also provide checklists for each layer and deliberate on potential sources of bias in each layer and their mitigation methodologies. This work will facilitate layer-wise standardisation of AI fairness rules and benchmarking parameters.
\end{abstract}


\section{Introduction}
\label{introduction}
The usage of Artificial Intelligence (AI) is widespread. Individuals use AI in recommendation systems, maps, health-monitoring apps, etc. They trust these applications and depend on them for several day-to-day decisions expecting them to be fair. Domain experts use AI to assist their work, such as the diagnostic and therapeutic tools used by doctors. Any bias in such systems can jeopardise the health of their patients and might lead to incorrect treatment or more visits to the hospital. Enterprises use AI for their internal systems, such as growth predictions, recruitment services, and fraud detection, besides the services they provide to their customers through their products. Biased systems would lead to a lack of credibility and business losses. Research organisations use AI to further their study and discover new areas in their field. Space exploration and medicinal research also use AI, where biased datasets may lead to the development of faulty products, increased budgets, and harmful consequences to society. Governments, judiciary, and defence services use AI for policy proposals, delivery of citizen-centric services, legal decisions, and warfare. Here, biased systems affect the rights of the citizens and can negatively affect the unprivileged class.

Therefore, bias mitigation and AI fairness are becoming ethical, social, commercial, and legal requirements. There is an urgent need to create a standardised holistic framework that addresses the bias concerns of various AI systems. The framework should include independent fairness ratings, transparency reports, and disclosures.

Data is the direct representation of society and the system that generates it. Several well-documented cases illustrate that society has not been fair to everyone historically. Even in today's times, discrimination is prevalent. When one trains AI systems using datasets representing such a society without proper checks, the systems are also highly likely to be biased. Many studies focus on identifying biases and mitigating them from the databases that can fall into the data pre-processing and model training phases of the lifecycle. But datasets are not the only source of bias, as other activities are also prone to biases. \citep{friedman2017bias} lists three categories of types of biases: preexisting, technical, and emergent. \citep{schwartz2022towards} categorises bias into systemic, statistical, and human biases.

Standard processes are required to mitigate bias and assess the fairness of a system in each phase of the AI development lifecycle. The challenges are multifold. AI fairness is context-sensitive. Developers use different types of machine learning (ML) techniques and algorithms for different types of AI systems. Even two similar types of systems require different approaches as each has a unique problem statement and a unique set of requirements. The consequences of biases vary for different scenarios. The sources and type of biases are also not the same for all applications. The tolerance for biases varies from case to case and also from region to region. Further, the concept of fairness is region and culture-dependent. Something considered fair in one country or culture may be regarded as biased in another. It is also data dependent. A face recognition model trained with European faces is more likely to be unbiased when deployed in France than in India.

Considering the varied requirements for each AI application, standardising a single end-to-end process for assessing fairness may not be feasible. However, most AI systems have similar development lifecycles, from framing the problem statement to their deployment and usage. Standardisation of bias mitigation procedures, fairness metrics, etc. is more pragmatic at each lifecycle stage, vaguely similar to the approach followed by the seven-layer Open System Interconnection (OSI) model for data communication. This paper proposes to handle fairness in a layered manner. For each lifecycle stage, we elaborate on the possibilities of biases, the precautions to avoid these biases, and the checklists to formalise the process.

This paper aims to empower designers, developers, evaluators, and service providers of AI applications to detect bias efficiently and develop fair AI systems. This paper not only helps standardise the fairness process but also makes it more thorough. It will also help individual developers, start-ups, and small organisations develop high-quality and fair AI systems competing with big organisations.

\subsection{Literature Review}
\label{literatureReview}
AI fairness is a rapidly growing research topic. In this section, we review the contemporary work related to biases observed in AI systems used by governments and leading corporations, biases due to imperfect datasets, biases due to algorithms, the concept of fairness, and metrics used to assess fairness. We also review works focusing on the typical AI lifecycle and standardising the bias mitigating processes.

\paragraph{AI Bias}
Instances of various deployed AI systems having biases based on demographics, such as race, gender, age, etc., are well documented. \citep{wadsworth2018achieving} cites racial bias in the AI system once used by the US judiciary for predicting the likelihood of a criminal committing a crime again. \citep{chowdhury2018auditing} reports that because of existing biases in the training data, several lenders in the US were granting loans to non-eligible white Americans while marking many eligible African Americans as ineligible. \citep{manrai2016genetic} demonstrates how the exclusion of African-Americans resulted in their misclassification in clinical studies. \citep{shankar2017no} analysed two of the most commonly used image datasets, ImageNet and Open Images, and found that the classifiers trained on these datasets gave varying results as the datasets were Amerocentric and Eurocentric. \citep{kodiyan2019overview} and others cite gender bias in the AI system used by Amazon to filter job applications as a result of training the AI system using data of historical hirings at Amazon when most of its employees were male. AI systems also exhibit age bias as training datasets often under-represent older people. In 2017, many hiring platforms were under investigation as they effectively excluded older applicants by not allowing them to enter the year of graduation before 1980 \citep{ajunwa2019beware}.

\citep{baeza2018bias} defines algorithm bias as bias that is not present in the data but added by the algorithm. \citep{akter2021addressing} conducts a thorough literature review focusing specifically on algorithm bias in AI and proposes two approaches to reduce algorithm bias in customer management: priori and post-hoc approaches. It also suggests that the stakeholders interact with the end-users and among themselves more to ensure the best-designed ethical AI system.

\citep{mehrabi2021survey} surveys different biases in machine learning and identifies different sources of biases. \citep{roselli2019managing} proposes a set of processes to manage and mitigate biases that may enter the AI system because of misrepresenting business goals in the AI algorithms, biased datasets, and inaccurate or stale data.

\paragraph{AI fairness definitions and metrics}
Different researchers define fairness in AI differently depending on the context of the problem statement. \citep{verma2018fairness} presents a compilation of various definitions of fairness proposed by researchers. \citep{mehrabi2021survey} categorises the fairness definitions into three types: individual fairness, group fairness, and subgroup fairness.

Fairness metrics help identify bias in AI systems and mathematically evaluate the fairness of such systems. \citep{castelnovo2022clarification} provides a detailed comparison of different fairness metrics and definitions. Commonly used fairness metrics include demographic parity, equal-opportunity \citep{hardt2016equality}, equal-mis-opportunity \citep{hinnefeld2018evaluating}, average odds, and distance metrics \citep{pandit2011comparative}. Each metric measures a different aspect of the data and the results. Demographic parity measures whether the probabilities of favourable outcomes for privileged and unprivileged groups of data are the same or not. Equal-opportunity makes sure that the model has the same recall for both groups.

\paragraph{AI lifecycle auditing}
Published work is also available on the AI lifecycle, though their focus is generally on the development lifecycle and not on fairness or bias mitigation. Researchers from Microsoft did a study to learn about the processes and practices followed by different teams at Microsoft while working on AI problems \citep{amershi2019software}. They used a nine-stage machine learning workflow and presented an ML process maturity metric to help teams self-assess how well they work with ML and to offer guidance for improvements. \citep{suresh2021framework} presents seven sources of bias in the seven phases of an ML lifecycle.

\citep{bantilan2018themis} proposed a Fairness Aware machine learning interface called themis-ml. The proposal has four interfaces each corresponding to the four components of the AI classification pipeline. The paper focuses only on simple binary classifiers and does not cover all the layers such as data ingestion.

\citep{raji2020closing} proposes an end-to-end auditing framework for developing AI systems to increase accountability. It suggests generating documents at each phase of the development lifecycle that are then internally audited and scrutinised.

\citep{mokander2022operationalising} presents a case study of AstraZeneca undergoing an ethics-based AI audit while discussing the limitations and challenges faced by organisations undergoing ethics-based AI audits.

\paragraph{AI ethics checklists and guidelines}
\citep{richardson2021framework} suggests software toolkits and checklists as the solutions to tackle algorithm bias and summarises recommendations for the normalisation of fairness practice. \citep{madaio2020co} highlights the importance of AI fairness checklists in organisations to formalise ad hoc processes and raises the need for the checklists to align with the existing workflows.

\citep{ryan2020artificial} provides a very comprehensive overview of various AI ethics guidelines.

\citep{hleg2020assessment} by High-Level Expert Group on Artificial Intelligence (AI HLEG) provides a detailed checklist of guidelines for the developers working on AI systems to self assess and ensure they develop along the principles of Trustworthy AI.

\citep{kumar2020trustworthy} proposes a framework for trustworthy AI systems focussing on the ethics of data and ethics of algorithms.

Our paper is a continuation of all these efforts by various researchers to identify biases and mitigate them to develop fair AI systems. Most fairness metrics use the predictions/ outputs of the AI model to evaluate its fairness. As a result, developers relying primarily on the fairness metrics learn about the model's fairness late in the development cycle, leading to more effort and cost. We provide a layered approach and checklists to tackle biases in the different phases of an AI system development lifecycle, thus allowing the stakeholders to take corrective measures from the beginning.

\subsection{Gap Analysis and Our Contribution}
\label{gapAnalysis}
Most researchers focus on a few aspects of fairness. We observe a gap in a holistic evaluation of fairness covering all activities from concept to usage of the AI system. Excessive focus on biases in the datasets, fairness metrics, and algorithmic bias leads to ignoring other aspects affecting AI fairness. Our paper attempts to fill this gap. To the best of our knowledge, we are the first to propose a seven-layer model to ensure fairness at every stage of the AI application lifecycle.

We propose a comprehensive approach in the form of a novel seven-layer model, inspired by the Open System Interconnection (OSI) model, to standardise AI fairness handling. Despite the differences in the various aspects, most AI systems have similar model-building stages. The proposed model splits the AI system lifecycle into seven abstraction layers, each corresponding to a well-defined AI model-building or usage stage. We also provide checklists for each layer and deliberate on potential sources of bias in each layer and their mitigation methodologies. This work will facilitate layer-wise standardisation of AI fairness rules and benchmarking parameters.

\subsection{Impact Statement}
\label{impactStatement}
Everyone from individuals to governments, startups to corporations use AI applications and expect them to be fair in their decisions. The users demand transparency. This requires standardisation of the fairness assessment and mitigation process. Considering the varied requirements, a single monolithic end-to-end standardised process may not be feasible. We propose to standardise the AI fairness assessment at functionally similar layers of the AI lifecycle. The seven-layer model along with fairness considerations and checklists for each layer proposed in this paper will help formalise the ad-hoc procedures that developers follow. Technically, it helps standardise the fairness assessment process. Socially, it helps build unbiased public-services applications. Legally, it helps develop a policy framework as the governments and regulators can specify layer-wise AI benchmarks. Agencies procuring AI applications can ask for layer-wise benchmarks and ratings as eligibility criteria. Economically, organisations can use the model to build fair applications and boost their businesses.

\section{The Seven-Layered Model}
\label{theSevenLayeredModel}
The proposed model consists of seven layers based on the AI lifecycle as follows:
\begin{enumerate}
\item Layer 1: Requirements, Context, and Purpose Layer
\item Layer 2: Data Collection and Selection Layer
\item Layer 3: Data Pre-processing and Feature Engineering Layer
\item Layer 4: Algorithm Layer
\item Layer 5: AI System Training Layer
\item Layer 6: Independent Audit Layer
\item Layer 7: Usage Layer
\end{enumerate}

\subsection{Layer 1: Requirements, Context, and Purpose Layer}
\label{layer1}
The first layer pertains to understanding the requirements, context, purpose, and concerns of the proposed AI system. This layer does not involve data handling or coding but is still essential in building a fair and unbiased AI system.

It involves understanding why AI is required to solve the problem statement. Knowing the current solution in place helps in better understanding the problem. The developers and the managers should brainstorm on the requirements. Knowledge of the domain allows the developers to grasp the possible biases they might encounter as they start development.

The scope and context of the AI system should be clear. There should be clear documentation of the expectations from the AI model, the intended users, and the anticipated deployment environment. The developers decide the machine learning model type best suited for the instant problem. They also identify if datasets are already available for the model training.

There should be due diligence for fairness. The service providers and the developers should identify the attributes that are prone to biases. There should be deliberation on how much is the tolerance/ acceptable limit for biases in the AI system. The demography of the users will lay the foundation for expected biases in the data and help the developers to take precautionary measures in the early stage of the development. Reviewing relevant AI literature helps in identifying any previously recorded instances of biases while solving similar problems.

This layer also involves identifying the AI fairness technique, such as group fairness or individual fairness, which will be used throughout the rest of the layers to identify biases in the AI model. There are multiple definitions of AI fairness, each having its own set of fairness measuring metrics. Identifying the technique at an early stage sets an effective measure to detect biases quickly.

Fairness checks in this layer:
\begin{enumerate}
\item What are the expectations from the AI model?
\item Are the requirements clear and formally documented?
\item Have the developers understood the scope and context of the problem?
\item How is the problem being dealt with presently?
\item Which fairness technique is appropriate and why?
\item How were the tolerance limits for fairness and bias decided?
\item Was the relevant AI literature reviewed?
\item Is the data generated internally or from external third-party/ public sources?
\item Were benchmarks and baselines identified?
\item Are Pre-trained models already available that can be repurposed?
\item What are the alternate solutions and flows?
\item Are the stakeholders, domain experts and auditors identified?
\end{enumerate}

\subsection{Layer 2: Data collection and selection Layer}
\label{layer2}
The data collection and selection layer comprises activities related to gathering data required for training the AI system. This layer includes identifying potential data sources, creating the data collection pipeline, and labelling the data based on the problem statement.

If the input data has biases, such as social biases, then the AI system trained on it will reflect the same biases. Thus, detecting and removing biases right at the initial stage is essential to ensure that the dataset is fair. After reviewing several AI-related biased outcomes, \citep{gupta2020algorithmic} concluded that biased training data is the foremost cause of bias in AI systems as the algorithms learn from the patterns in the training dataset.

The data collection process may introduce bias, even though the dataset is not biased. Often, data is either not labelled or needs to be labelled differently based on the context of the problem statement. In such situations, the individual developers working on the AI system label the data. Companies frequently outsource the labelling job to a third party, such as Amazon Mechanical Turk \citep{sorokin2008utility}. This has a high chance of introducing bias in the system, as the understanding and personal perceptions of the individuals labelling the data decide the labels assigned to the data. A recent MIT study \citep{northcutt2021pervasive} found thousands of mislabeled samples in the datasets used to train commercial systems.

It could be possible that the data collected is not balanced because of selective filtering/ sampling of data. \citep{buolamwini2018gender} suggests that sampling bias can cause poor generalisation of the learning algorithms. Developers might leave out important data because of either their prejudice or ignorance of its importance.

A critical issue is unbalanced data at the source. People belonging to a certain race, gender, or community are often under-represented, as compared to others, in various datasets because of historical discrimination or restricted access to services. This disparity in data quality is also a cause of algorithm bias \citep{fu2020ai}. Also, data of certain groups would be incomplete or inaccurate compared to other groups, leading to more noise in the dataset for such groups.

Too much data is also harmful. AI systems can make better predictions with less information too. The fundamental point is that whatever information is available is relevant to the problem statement. To increase the training dataset size, developers sometimes add extra information that might not apply to the instant case, leading to information bias. Also, to expand the dataset size, data might be collected from unverified sources, thus raising the risk of bias in the resultant AI system.

Data captured using defective or faulty devices can lead to measurement/ device bias \citep{srinivasan2021biases}. For example, a poorly calibrated Internet of Things (IoT) sensor can lead to false alerts, and a system trained using its data will be biased and inaccurate.

Thus, systematic selection of appropriate variables and collection of relevant data is necessary to reduce algorithm bias \citep{richardson2021framework}.

We suggest various checks to ensure that the data collected and selected for system building is fair and unbiased.

Fairness checks while selecting a dataset:
\begin{enumerate}
\item Is the dataset collected from a known and verifiable data repository?
\item Is there a way to audit the dataset for validity and accuracy?
\item Is the dataset complete? Are there too many missing values?
\item Is the dataset cleaned and pre-processed or raw?
\item How distributed is the dataset? Does it include data from all sections of society? Is it skewed in favour of certain groups?
\item Is there any metadata available along with the dataset that increases its explainability? Is the information available around who collected it, when it was collected, and the collection processes used?
\item Why was this data collected?
\item How old is the dataset? Do the situations/ assumptions at the time of data collection still apply?
\item Is more than one dataset being considered? If yes, what is the correlation between them?
\end{enumerate}

Fairness checks while collecting and labelling data:
\begin{enumerate}
\item Is the dataset already labelled?
\item In the case of manual labelling, are the people labelling the data trained and well aware of the context of the problem?
\item Are quality assurance and verification processes included to ensure no developer bias gets into the dataset?
\end{enumerate}

\subsection{Layer 3: Data Pre-processing and Feature Engineering Layer}
\label{layer3}
Data pre-processing and feature engineering form the next layer. Most of the datasets used for training the model need pre-processing. The same dataset can train multiple AI systems, each having a different problem statement, requiring different input data formats, and/ or requiring separate labels. Some AI systems might require aggregated data, while others might work on a different numeric scale altogether. Some AI systems might need only a subset of records or use only a few features.

Pre-processing is the set of steps required to make the raw data suitable for analysis and training. It primarily involves assessing data quality, cleaning the data, and data transformation. Depending on the problem statement and the dataset quality, one might have to remove outliers, correct data type mismatches, remove null values or fill them with synthetic values, and perform normalisation, among other pre-processing steps.

Bias can easily creep into the system in this layer. Aggregating and normalising the records can lead to measurement and aggregation biases. Important features might get sidelined or scaled down to a level that they lose significance, leading to omitted variable bias \citep{ayres2010testing} and exclusion bias. Too much synthetic data is not a good representation of the ground reality and can lead to unfair predictions by the AI system. Removing rows with null values or outliers can lead to a significant drop in the number of records of a particular demography/ section of society that, because of historical reasons and discrimination, might not have had that much access, leading to fewer and improper records.

The identification of protected attributes and privileged classes for each protected attribute is the first step toward developing a fair AI system. Protected attributes are usually those features that are associated with social bias. Some common protected attributes are gender, race, caste, religion, etc. Protected attributes are application dependent; a protected attribute for one application may not be so for another. For example, gender may be a protected attribute for a recruitment application but not for a medical application, such as breast cancer patient forecasting. Protected attributes should be decided meticulously and ensured that the developer's personal bias does not play into the picture.

Another component of pre-processing is bias mitigation. If there are already known biases in the dataset, we use mitigation techniques to reduce their influence on model training to a certain extent. Techniques such as sampling, reweighing, and data transformation have proved useful in reducing the effect of biases in the dataset \citep{kamiran2012data, calmon2017optimized}.

Fairness checks in this layer:
\begin{enumerate}
\item What are the approaches taken to clean the dataset? Are they leading to an imbalanced dataset?
\item How is feature engineering done? Was it done by a single developer or reviewed by others also?
\item Were any features dropped? If yes, why and what effect do they have on the problem?
\item Would the pre-processing process remove an enormous chunk of data from particular demography?
\item As part of over-sampling, did the algorithm add too much synthetic data?
\item How were the normalisation and scaling techniques decided?
\item How were the protected attributes and privileged classes selected?
\end{enumerate}

\subsection{Layer 4: Algorithm Layer}
\label{layer4}
The Algorithm layer is a set of activities related to writing the actual code for the model logic. It involves selecting the most appropriate machine learning type,  algorithm, and fairness criteria and writing the code.

This layer is susceptible to developer bias at several junctures. All developers perceive fairness in their way. If not checked, their individual beliefs will affect the overall output of the models they build. As a solution, companies are increasingly using automated tools to make accurate and comprehensive choices. However, as \citep{polli2017dark} noted, while reliance on technology may avoid human bias, the potential to produce biased algorithms opens up a dark side of algorithm-based decisions.

The machine learning technique used, out of supervised, unsupervised, semi-supervised, or reinforcement learning, has an enormous impact on the overall fairness of the AI system. For a given case, all the ML techniques are not equally appropriate. Some provide good results in one situation but lead to over-fitting in another. Over-fitting around the privileged class leads to an unfair AI system that gives biased results against the unprivileged class. The Robodebt scheme in Australia wrongly and unlawfully pursued hundreds of thousands of welfare clients for the debt they did not owe \citep{whiteford2021debt, akter2021addressing}. Biased vehicular automation may cause severe damage, as evident in the accidents caused by the self-driving cars by UBER \citep{wakabayashi2018self} or in the incident of deaths caused by the malfunctioned robot at an Amazon warehouse \citep{shah2018amazon}.

Many algorithms involve auto-encoders that reduce features of the dataset so that it becomes easier to learn. This feature extraction process can introduce bias as it might discard contextual information that could be relevant in some situations \citep{siwicki2021how}.

The black-box nature of AI algorithms makes it difficult to debug the root cause behind bias in the AI system. Developers should attempt to make the AI system development process as transparent and explainable as possible. Model interpretability should be a factor when selecting an algorithm. One should always prefer simple models over highly complex black-box models. \citep{angelino2018learning} and other studies observe that the performance of several simple models is as good and accurate as the state-of-the-art models.

This layer also includes the identification of the preferred choice of fairness technique. There are multiple ways to define and analyze the AI system fairness, such as individual fairness and group fairness \citep{aggarwal2019black}.

Fairness checks in the development phase:
\begin{enumerate}
\item Is the development process transparent?
\item Are there regular reviews and quality assessments to make sure the developer's individual biases are not affecting the model?
\item What were the criteria for selecting the algorithm technique?
\item Were the stakeholders consulted? Was the context of the problem statement taken into consideration?
\item Which fairness technique was selected - individual fairness or group fairness, or something else?
\item Is the model developed per the requirements specified in layer 1?
\end{enumerate}

\subsection{Layer 5: AI System Training Layer}
\label{layer5}
The AI system training layer relates to activities associated with the ML model training. This stage includes dividing the dataset into training, testing, and validation sets, training the model iteratively, and evaluating its fairness using fairness metrics.

The model trains on the training dataset through multiple iterations. After each iteration, the developers evaluate its performance using the testing dataset and determine how many iterations are required. Finally, they use the validation dataset to determine the expected performance of the model in the field. Thus, distributing data in the three datasets without prejudice is crucial for the model's quality. Even if the input dataset is unbiased, its biased or inappropriate splitting into the three datasets could lead to biases in the AI system. While a biased training dataset leads to a bias in the trained model, a biased testing or validation dataset does not fairly evaluate the performance of the trained model.

One of the steps in model training is hyper-tuning. Hyper-tuning of parameters is prone to developer bias. Different developers interpret the results differently. They might prefer one optimization function over another or select a different learning rate. All these decisions decide how the hyper-parameters are tuned.

For the same problem statement, different developers might consider separate parameters to represent the model performance. It could be accuracy, precision, recall, f1 score, or something else. There could be scenarios where recall is more important than precision or vice versa. For example, if the aim is to select and publish the top ten positive user reviews for a hotel, the model should detect all positive reviews. It may mark some negative reviews as positive with a low score (i.e. higher recall) instead of leaving out some positive reviews to avoid some negative ones getting marked as positive (i.e. higher precision). Thus, correct performance parameter selection is essential and directly impacts the model’s fairness.

An overfit or an underfit model will likely be unfair and biased towards one data section. The trade-off between fairness and performance also reflects individual and organisation bias.

Selecting inappropriate benchmarks for comparing the model performance may lead to evaluation bias. For example, some benchmarks used to test facial recognition systems, such as Adience and IJB-A, are known to be biased toward skin color and gender \citep{buolamwini2018gender}.

After training the model, check fairness for each protected attribute. One should evaluate the fairness of both the training dataset as well as the model outcomes for the test dataset. Selecting the most appropriate fairness metrics for each protected attribute is crucial. The developers may use multiple fairness metrics for each attribute. Some commonly used fairness metrics are statistical parity difference, disparate impact, equal opportunity, equal mis-opportunity, and average odds.

Fairness check in the training phase:
\begin{enumerate}
\item How was the dataset divided into training, testing, and validation sets? Was the data distributed to them randomly?
\item Were the resultant three datasets balanced? Did they have an equal/ proportional distribution of privileged and unprivileged classes?
\item How were the hyper-parameters selected?
\item What are the fairness metrics used, and why?
\item Is fairness checked for each protected attribute or not?
\item What are the identified benchmarks, parameters, and metrics for measuring model performance, and why?
\item The choice of metrics, parameters, and benchmarks should be well documented and available for external auditing and explainability.
\end{enumerate}

\subsection{Layer 6: Independent Audit Layer}
\label{layer6}
The Independent Audit layer includes testing and fairness certification by a team other than the developers. It requires a standardised fairness assessment process, a rating metric independent of the various fairness metrics used in layer 5, and independent auditors.

With the increasing use of AI in all aspects of our lives, the users want to know that the decisions that the AI systems make are fair or not \citep{agarwal2022fairness}. Demand for independent audits based on a standardised process and certification or rating of AI systems before deploying them is fast increasing. A standard certification process is required to increase trust among the users of the AI model.

Various governments are using AI applications to deliver public services and identify beneficiaries of social welfare schemes. Such applications must not be biased against or in favour of some groups of citizens. The judiciary is fast adopting AI systems to assist in court cases. Investigation agencies use AI to monitor online crime, facial recognition, and threat prediction, among many other use cases. Smart cities, traffic control, banking, and many other public services are incorporating AI systems at various levels to provide better facilities. With the increasing convergence of AI and public services, AI fairness certification would soon be a legal requirement, not just an ethical concern.

\citep{kazim2021systematizing} examines the need for AI audit with a specific focus on AI used in recruitment systems. \citep{landers2022auditing} discusses that psychological audits can evaluate the fairness and bias of AI systems that make predictions about humans across disciplinary perspectives.

Independent audit and certification also help in increasing the model's explainability and provide transparency. Additionally, the service providers get another factor to consider while comparing and selecting AI models for their services. Governments and multilateral agencies following a transparent procurement process could use the fairness certificate as an eligibility criterion.

\citep{agarwal2022fairness} proposes a standardised process for assessing the fairness of supervised-learning AI systems and rating them on a linear scale. Accredited third-party agencies or a separate in-house team not involved in the model building may do the audit following the standardised process.

The auditor should evaluate fairness for each protected attribute. Fairness metrics used may vary as per attributes and also use cases. Uniformity and comparison between different AI models require a universal standard rating metric. This metric should derive from the individual fairness metrics used in layer 5, but its score should be independent of such metrics. It should also give a single score for the overall model instead of separate scores for each protected attribute. Fairness Score and Bias Index proposed by \citep{agarwal2022fairness} are appropriate for measuring the overall fairness of the AI system and the bias for each protected attribute.

Fairness checks in this layer:
\begin{enumerate}
\item Is the auditor independent? If it is an in-house team, then is it different from the developers? If it is a third-party auditor, then is it accredited?
\item Is a standard fairness assessment process followed?
\item Is the rating matric universal?
\item Has the auditor checked all the protected attributes?
\end{enumerate}

\subsection{Layer 7: Usage Layer}
\label{layer7}
The topmost layer is the Usage Layer. Here, the AI system moves from development and testing to commercial use. This layer involves deploying the AI system in the field, its regular monitoring and periodic retraining, and timely maintenance.

Deploying an AI system designed and trained for a particular situation in a different setting could adversely affect not only its business but also its fairness. AI systems trained for facial recognition with the dataset of Asian faces might not be accurate and fair if deployed in Europe or the USA.

Data is dynamic; the ground reality keeps on changing. A dataset used in training the system may, after some time, no longer represent the prevailing situations. It is also possible that the business goals shift or there is a change in the underlying technology. All these factors might cause an AI model, fair at the time of deployment, to give biased results.

There is a need for regularly retraining the model while accounting for the changing situations. The retraining keeps the model updated with the latest trends, leading to better accuracy and fairness. There are two ways to use the new data gathered since the last system training for the AI system retraining. One way is to add the new data to the old data and retrain the system on the complete data. The other way is to discard the old data and use only the newly gathered data for system retraining. One can select either of the two approaches based on the use case, data generation speed, and the time gap between the two training sessions.

Before retaining, we should perform fairness checks on the new data as a skewed use of the AI system may lead to biases in such new data.

After deploying the system, regularly monitor its performance using the fairness metrics used in the previous layers as Key Performance Indicators (KPIs). Any drop in the metrics should be an early sign of the need for system retraining.

Fairness checks in this layer:
\begin{enumerate}
\item Are fairness KPIs regularly monitored after the release?
\item How much time has passed since the last round of model training?
\item Have the underlying ground realities changed since deploying the model?
\item Is the distribution of protected classes in the protected attributes the same in production data as in the training data?
\item What were the deciding factors for retaining or discarding the old data while retraining the model?
\end{enumerate}

\section{Case study and discussion}
\label{discussion}
As a case study, the AI system to predict the risk of loan repayment analysed by \citep{agarwal2022fairness} was passed through the seven-layer model. It uses the German Credit Dataset \citep{dua2019} for training and validation. The application is expected to be used by users in an Asian country.

Layer 1 checklist indicates that the context is different, and the social and economic conditions of the target audience are very different from those available in the training dataset. Some additional demographic details should be treated as protected attributes but not available in the dataset.

Layer 2 checks indicate that the dataset is quite old and may not be a true representative of the prevailing scenario.

The model building involved pre-processing (layer 3), using logistic regression (layer 4), and randomly dividing the dataset into training and testing datasets (layer 5). The application trained with the original dataset showed some bias on the protected attribute “Sex” with the values of the fairness metrics Disparate Impact and Equal Mis-Opportunity Difference being 0.8816 and 0.1333, respectively.

Layer 6 analysis shows that the Bias Index and Fairness Score were 0.1037 and 0.8963, which after mitigating bias in the protected attribute, changed to 0.0864 and 0.9136, respectively.

The checklist of layer 7 indicates usage other than the intended one.

While the fairness metrics and the overall Fairness Score give a misleading impression that the application is fair, the checks performed in layers 1, 2 and 7 show that the application may not be fair for the target users. With readymade applications increasingly available, there are growing chances of using the application for demography other than for which it was developed. The case study confirms that the seven-layered model and checklists suggested for each level will ensure that the application is fair for the intended use case and that the developers identify the requirements and challenges at the initial stages itself.

The role of the seven layers are summarised as follows:

Layer 1: Bias mitigation starts at the first stage itself with due diligence about AI fairness. The developers should clearly understand the scope of the problem statement, the context, and the expectations from the AI system before the actual development starts. Identifying data sources, protected attributes, possible privileged and unprivileged classes, ML techniques, fairness models, and understanding tolerance limits to bias helps build a fairer AI system.

Layer 2: The most common source of bias entering the AI system is the datasets used in the model training. Biases can be avoided by selecting datasets only after verifying the data source and auditing the datasets for fairness. Selecting updated datasets will ensure that the AI model trains with the latest trends and does not learn the defunct prejudices that were once prominent in society.

Layer 3: Effective pre-processing and feature engineering not only improves the model accuracy but also reduces biases in the datasets. Bias mitigation techniques such as oversampling and adversarial debiasing have proved successful. Formally define the protected attributes and privileged classes.

Layer 4: Developer bias at the algorithm layer may be mitigated by engaging a diverse team of developers, not relying on an individual, and getting the code reviewed by another team. Confirm that the selected ML technique and algorithms are appropriate for the given case. Use simple and explainable models.

Layer 5: Splitting the input dataset into training, testing, and validation datasets in an unbiased manner, so that each dataset properly represents the target audience is necessary to ensure fair model building. Selecting the appropriate fairness metrics while iteratively training and testing the model is crucial to ascertain that the model fairness is within the tolerance limits as decided in layer 1.

Layer 6: Demand for independent audits will rapidly grow as governments and the judiciary increasingly use AI systems for citizen-centric services. A standardised fairness assessment process and a universal rating metric, as proposed by \citep{agarwal2022fairness}, are pivotal for this layer. Self-disclosures of the fairness ratings and ecosystem of accredited certifying agencies will propel awareness about AI fairness and result in better compliance with fairness requirements.

Layer 7: Biases during usage can be mitigated by ensuring usage as per design assumptions, regularly checking the fairness KPIs, retraining the system whenever required, and mitigating bias from the newly gathered data before its use for retraining.

\section{Conclusion and Future Work}
\label{conclusion}
AI systems are susceptible to biases at every stage, from conceiving the project to its usage. This paper proposes to standardise the bias detection and mitigation process throughout the AI system lifecycle through a novel seven-layered model. Conventional approaches for tackling biases usually focus more on a few specific aspects of the AI systems, with the possibility of other elements getting ignored. A layered approach, such as the one proposed in this paper, enables a holistic view of bias handling.

The seven-layer model proposed in this paper provides a more structured, systematic, and standard approach to dealing with AI biases, required as we move from the present-day soft-touch self-regulatory conditions to a more regulated environment for AI fairness. It enables independent fairness assessment at each layer and facilitates process standardisation. The checklists proposed for each layer formalise the ad-hoc processes presently followed to check for fairness. 

With the growing use of AI applications in all domains and increasing awareness of the need for fairness, customers will demand more disclosures on the fairness benchmarks. Also, governments and regulatory bodies are actively pursuing policy formulations and regulations to ensure AI fairness. This work will help individual developers, start-ups, and small organisations that lack institutional procedures develop fair AI systems. It will also help governments and sectoral regulatory bodies to notify layer-wise fairness benchmarks for various critical applications.

The paper proposes to split the AI system lifecycle into seven distinct layers - requirements, context and purpose, data collection and selection, data pre-processing and feature engineering, algorithm building, AI system training, independent audit and fairness certification, and usage. Layers separate from each other based on the activities involved. While the layers are interdependent, one can view each layer independently for bias handling. An analogy with the OSI framework helps to understand the concept of layers in the context of fairness. The paper also elaborates on the roles in each layer of various stakeholders, such as developers, service providers, and users.

The paper also discusses the various types of biases possible at each layer and the questions to be asked during model building to avoid these biases. Each layer uses different metrics and benchmarks to assess and measure fairness. Layer 5 (AI system training) uses fairness metrics such as disparate impact, equal opportunity difference, statistical parity difference, etc. Layer 6 (Independent audit and fairness certification) uses rating benchmarks such as Fairness Score and Bias Index.

AI fairness is context-sensitive and case-specific. A single framework for assessing fairness for all AI systems is not pragmatic. Still, various types of AI systems have similar stages in model building. The layered approach provides abstraction at each functional layer, thus facilitating the analysis and comparison of fairness requirements for different types of AI systems at the layer level. The AI system is fair if all the layers meet the layer-specific fairness requirements.

To fully implement the concept of layered handling, we suggest further studies for formulating fairness benchmarks, akin to the Fairness Score, for all layers. The ethical AI rules, assessment procedures, and fairness measurement metrics can be defined for each layer, thus facilitating standardisation.

\section{Declarations}
\label{declaration}
\textbf{Funding} The authors did not receive support from any organization for the submitted work. \\
\textbf{Availability of data and material} Not applicable. \\
\textbf{Conflicts of interest/ Competing interests} On behalf of all authors, the corresponding author states that there is no conflict of interest. \\
\textbf{Code availability} Not applicable. \\
\textbf{Disclamer} The views or opinions expressed in this paper are solely those of the authors and do not necessarily represent those of their respective organizations.

\bibliographystyle{unsrtnat}

\begin{thebibliography}{47}
\providecommand{\natexlab}[1]{#1}
\providecommand{\url}[1]{\texttt{#1}}
\expandafter\ifx\csname urlstyle\endcsname\relax
  \providecommand{\doi}[1]{doi: #1}\else
  \providecommand{\doi}{doi: \begingroup \urlstyle{rm}\Url}\fi

\bibitem[Friedman and Nissenbaum(2017)]{friedman2017bias}
Batya Friedman and Helen Nissenbaum.
\newblock Bias in computer systems.
\newblock In \emph{Computer Ethics}, pages 215--232. Routledge, 2017.

\bibitem[Schwartz et~al.(2022)Schwartz, Vassilev, Greene, Perine, Burt, Hall,
  et~al.]{schwartz2022towards}
Reva Schwartz, Apostol Vassilev, Kristen Greene, Lori Perine, Andrew Burt,
  Patrick Hall, et~al.
\newblock Towards a standard for identifying and managing bias in artificial
  intelligence.
\newblock \emph{Special Publication (NIST SP), National Institute of Standards
  and Technology}, 2022.

\bibitem[Wadsworth et~al.(2018)Wadsworth, Vera, and
  Piech]{wadsworth2018achieving}
Christina Wadsworth, Francesca Vera, and Chris Piech.
\newblock Achieving fairness through adversarial learning: an application to
  recidivism prediction.
\newblock \emph{arXiv preprint arXiv:1807.00199}, 2018.

\bibitem[Chowdhury and Mulani(2018)]{chowdhury2018auditing}
Rumman Chowdhury and Narendra Mulani.
\newblock Auditing algorithms for bias.
\newblock \emph{Harvard Business Review}, 24, 2018.

\bibitem[Manrai et~al.(2016)Manrai, Funke, Rehm, Olesen, Maron, Szolovits,
  Margulies, Loscalzo, and Kohane]{manrai2016genetic}
Arjun~K Manrai, Birgit~H Funke, Heidi~L Rehm, Morten~S Olesen, Bradley~A Maron,
  Peter Szolovits, David~M Margulies, Joseph Loscalzo, and Isaac~S Kohane.
\newblock Genetic misdiagnoses and the potential for health disparities.
\newblock \emph{New England Journal of Medicine}, 375\penalty0 (7):\penalty0
  655--665, 2016.

\bibitem[Shankar et~al.(2017)Shankar, Halpern, Breck, Atwood, Wilson, and
  Sculley]{shankar2017no}
Shreya Shankar, Yoni Halpern, Eric Breck, James Atwood, Jimbo Wilson, and
  D~Sculley.
\newblock No classification without representation: Assessing geodiversity
  issues in open data sets for the developing world.
\newblock \emph{arXiv preprint arXiv:1711.08536}, 2017.

\bibitem[Kodiyan(2019)]{kodiyan2019overview}
Akhil~Alfons Kodiyan.
\newblock An overview of ethical issues in using ai systems in hiring with a
  case study of amazon’s ai based hiring tool.
\newblock \emph{Researchgate Preprint}, pages 1--19, 2019.

\bibitem[Ajunwa(2019)]{ajunwa2019beware}
Ifeoma Ajunwa.
\newblock Beware of automated hiring.
\newblock \emph{The New York Times}, 8, 2019.

\bibitem[Baeza-Yates(2018)]{baeza2018bias}
Ricardo Baeza-Yates.
\newblock Bias on the web.
\newblock \emph{Communications of the ACM}, 61\penalty0 (6):\penalty0 54--61,
  2018.

\bibitem[Akter et~al.(2021)Akter, Dwivedi, Biswas, Michael, Bandara, and
  Sajib]{akter2021addressing}
Shahriar Akter, Yogesh~K Dwivedi, Kumar Biswas, Katina Michael, Ruwan~J
  Bandara, and Shahriar Sajib.
\newblock Addressing algorithmic bias in ai-driven customer management.
\newblock \emph{Journal of Global Information Management (JGIM)}, 29\penalty0
  (6):\penalty0 1--27, 2021.

\bibitem[Mehrabi et~al.(2021)Mehrabi, Morstatter, Saxena, Lerman, and
  Galstyan]{mehrabi2021survey}
Ninareh Mehrabi, Fred Morstatter, Nripsuta Saxena, Kristina Lerman, and Aram
  Galstyan.
\newblock A survey on bias and fairness in machine learning.
\newblock \emph{ACM Computing Surveys (CSUR)}, 54\penalty0 (6):\penalty0 1--35,
  2021.

\bibitem[Roselli et~al.(2019)Roselli, Matthews, and
  Talagala]{roselli2019managing}
Drew Roselli, Jeanna Matthews, and Nisha Talagala.
\newblock Managing bias in ai.
\newblock In \emph{Companion Proceedings of The 2019 World Wide Web
  Conference}, pages 539--544, 2019.

\bibitem[Verma and Rubin(2018)]{verma2018fairness}
Sahil Verma and Julia Rubin.
\newblock Fairness definitions explained.
\newblock In \emph{2018 ieee/acm international workshop on software fairness
  (fairware)}, pages 1--7. IEEE, 2018.

\bibitem[Castelnovo et~al.(2022)Castelnovo, Crupi, Greco, Regoli, Penco, and
  Cosentini]{castelnovo2022clarification}
Alessandro Castelnovo, Riccardo Crupi, Greta Greco, Daniele Regoli,
  Ilaria~Giuseppina Penco, and Andrea~Claudio Cosentini.
\newblock A clarification of the nuances in the fairness metrics landscape.
\newblock \emph{Scientific Reports}, 12\penalty0 (1):\penalty0 1--21, 2022.

\bibitem[Hardt et~al.(2016)Hardt, Price, and Srebro]{hardt2016equality}
Moritz Hardt, Eric Price, and Nati Srebro.
\newblock Equality of opportunity in supervised learning.
\newblock \emph{Advances in neural information processing systems}, 29, 2016.

\bibitem[Hinnefeld et~al.(2018)Hinnefeld, Cooman, Mammo, and
  Deese]{hinnefeld2018evaluating}
J~Henry Hinnefeld, Peter Cooman, Nat Mammo, and Rupert Deese.
\newblock Evaluating fairness metrics in the presence of dataset bias.
\newblock \emph{arXiv preprint arXiv:1809.09245}, 2018.

\bibitem[Pandit et~al.(2011)Pandit, Gupta, et~al.]{pandit2011comparative}
Shraddha Pandit, Suchita Gupta, et~al.
\newblock A comparative study on distance measuring approaches for clustering.
\newblock \emph{International journal of research in computer science},
  2\penalty0 (1):\penalty0 29--31, 2011.

\bibitem[Amershi et~al.(2019)Amershi, Begel, Bird, DeLine, Gall, Kamar,
  Nagappan, Nushi, and Zimmermann]{amershi2019software}
Saleema Amershi, Andrew Begel, Christian Bird, Robert DeLine, Harald Gall, Ece
  Kamar, Nachiappan Nagappan, Besmira Nushi, and Thomas Zimmermann.
\newblock Software engineering for machine learning: A case study.
\newblock In \emph{2019 IEEE/ACM 41st International Conference on Software
  Engineering: Software Engineering in Practice (ICSE-SEIP)}, pages 291--300.
  IEEE, 2019.

\bibitem[Suresh and Guttag(2021)]{suresh2021framework}
Harini Suresh and John Guttag.
\newblock A framework for understanding sources of harm throughout the machine
  learning life cycle.
\newblock In \emph{Equity and access in algorithms, mechanisms, and
  optimization}, pages 1--9. 2021.

\bibitem[Bantilan(2018)]{bantilan2018themis}
Niels Bantilan.
\newblock Themis-ml: A fairness-aware machine learning interface for end-to-end
  discrimination discovery and mitigation.
\newblock \emph{Journal of Technology in Human Services}, 36\penalty0
  (1):\penalty0 15--30, 2018.

\bibitem[Raji et~al.(2020)Raji, Smart, White, Mitchell, Gebru, Hutchinson,
  Smith-Loud, Theron, and Barnes]{raji2020closing}
Inioluwa~Deborah Raji, Andrew Smart, Rebecca~N White, Margaret Mitchell, Timnit
  Gebru, Ben Hutchinson, Jamila Smith-Loud, Daniel Theron, and Parker Barnes.
\newblock Closing the ai accountability gap: Defining an end-to-end framework
  for internal algorithmic auditing.
\newblock In \emph{Proceedings of the 2020 conference on fairness,
  accountability, and transparency}, pages 33--44, 2020.

\bibitem[M{\"o}kander and Floridi(2022)]{mokander2022operationalising}
Jakob M{\"o}kander and Luciano Floridi.
\newblock Operationalising ai governance through ethics-based auditing: an
  industry case study.
\newblock \emph{AI and Ethics}, pages 1--18, 2022.

\bibitem[Richardson and Gilbert(2021)]{richardson2021framework}
Brianna Richardson and Juan~E Gilbert.
\newblock A framework for fairness: A systematic review of existing fair ai
  solutions.
\newblock \emph{arXiv preprint arXiv:2112.05700}, 2021.

\bibitem[Madaio et~al.(2020)Madaio, Stark, Wortman~Vaughan, and
  Wallach]{madaio2020co}
Michael~A Madaio, Luke Stark, Jennifer Wortman~Vaughan, and Hanna Wallach.
\newblock Co-designing checklists to understand organizational challenges and
  opportunities around fairness in ai.
\newblock In \emph{Proceedings of the 2020 CHI Conference on Human Factors in
  Computing Systems}, pages 1--14, 2020.

\bibitem[Ryan and Stahl(2020)]{ryan2020artificial}
Mark Ryan and Bernd~Carsten Stahl.
\newblock Artificial intelligence ethics guidelines for developers and users:
  clarifying their content and normative implications.
\newblock \emph{Journal of Information, Communication and Ethics in Society},
  2020.

\bibitem[HLEG(2020)]{hleg2020assessment}
AI~HLEG.
\newblock Assessment list for trustworthy artificial intelligence (altai) for
  self-assessment.
\newblock \emph{High Level Expert Group on Artificial Intelligence. B-1049
  Brussels}, 2020.

\bibitem[Kumar et~al.(2020)Kumar, Braud, Tarkoma, and
  Hui]{kumar2020trustworthy}
Abhishek Kumar, Tristan Braud, Sasu Tarkoma, and Pan Hui.
\newblock Trustworthy ai in the age of pervasive computing and big data.
\newblock In \emph{2020 IEEE International Conference on Pervasive Computing
  and Communications Workshops (PerCom Workshops)}, pages 1--6. IEEE, 2020.

\bibitem[Gupta and Krishnan(2020)]{gupta2020algorithmic}
Damini Gupta and TS~Krishnan.
\newblock Algorithmic bias: Why bother.
\newblock \emph{California Management Review}, 63\penalty0 (3), 2020.

\bibitem[Sorokin and Forsyth(2008)]{sorokin2008utility}
Alexander Sorokin and David Forsyth.
\newblock Utility data annotation with amazon mechanical turk.
\newblock In \emph{2008 IEEE computer society conference on computer vision and
  pattern recognition workshops}, pages 1--8. IEEE, 2008.

\bibitem[Northcutt et~al.(2021)Northcutt, Athalye, and
  Mueller]{northcutt2021pervasive}
Curtis~G Northcutt, Anish Athalye, and Jonas Mueller.
\newblock Pervasive label errors in test sets destabilize machine learning
  benchmarks.
\newblock \emph{arXiv preprint arXiv:2103.14749}, 2021.

\bibitem[Buolamwini and Gebru(2018)]{buolamwini2018gender}
Joy Buolamwini and Timnit Gebru.
\newblock Gender shades: Intersectional accuracy disparities in commercial
  gender classification.
\newblock In \emph{Conference on fairness, accountability and transparency},
  pages 77--91. PMLR, 2018.

\bibitem[Fu et~al.(2020)Fu, Huang, and Singh]{fu2020ai}
Runshan Fu, Yan Huang, and Param~Vir Singh.
\newblock Ai and algorithmic bias: Source, detection, mitigation and
  implications.
\newblock \emph{Detection, Mitigation and Implications (July 26, 2020)}, 2020.

\bibitem[Srinivasan and Chander(2021)]{srinivasan2021biases}
Ramya Srinivasan and Ajay Chander.
\newblock Biases in ai systems.
\newblock \emph{Communications of the ACM}, 64\penalty0 (8):\penalty0 44--49,
  2021.

\bibitem[Ayres(2010)]{ayres2010testing}
Ian Ayres.
\newblock Testing for discrimination and the problem of” included variable
  bias”,”.
\newblock \emph{Yale Law School Mimeo}, 2010.

\bibitem[Kamiran and Calders(2012)]{kamiran2012data}
Faisal Kamiran and Toon Calders.
\newblock Data preprocessing techniques for classification without
  discrimination.
\newblock \emph{Knowledge and information systems}, 33\penalty0 (1):\penalty0
  1--33, 2012.

\bibitem[Calmon et~al.(2017)Calmon, Wei, Vinzamuri, Natesan~Ramamurthy, and
  Varshney]{calmon2017optimized}
Flavio Calmon, Dennis Wei, Bhanukiran Vinzamuri, Karthikeyan
  Natesan~Ramamurthy, and Kush~R Varshney.
\newblock Optimized pre-processing for discrimination prevention.
\newblock \emph{Advances in neural information processing systems}, 30, 2017.

\bibitem[Polli(2017)]{polli2017dark}
Frida Polli.
\newblock The dark side of artificial intelligence, 2017.

\bibitem[Whiteford(2021)]{whiteford2021debt}
Peter Whiteford.
\newblock Debt by design: The anatomy of a social policy fiasco--or was it
  something worse?
\newblock \emph{Australian Journal of Public Administration}, 80\penalty0
  (2):\penalty0 340--360, 2021.

\bibitem[Wakabayashi(2018)]{wakabayashi2018self}
Daisuke Wakabayashi.
\newblock Self-driving uber car kills pedestrian in arizona, where robots roam.
\newblock \emph{The New York Times}, 19\penalty0 (03), 2018.

\bibitem[Shah(2018)]{shah2018amazon}
Saqib Shah.
\newblock Amazon workers hospitalized after warehouse robot releases bear
  repellent, 2018.

\bibitem[Siwicki(2021)]{siwicki2021how}
Bill Siwicki.
\newblock How ai bias happens – and how to eliminate it, 2021.

\bibitem[Angelino et~al.(2018)Angelino, Larus-Stone, Alabi, Seltzer, and
  Rudin]{angelino2018learning}
Elaine Angelino, Nicholas Larus-Stone, Daniel Alabi, Margo Seltzer, and Cynthia
  Rudin.
\newblock Learning certifiably optimal rule lists for categorical data.
\newblock \emph{Journal of Machine Learning Research}, 18:\penalty0 1--78,
  2018.

\bibitem[Aggarwal et~al.(2019)Aggarwal, Lohia, Nagar, Dey, and
  Saha]{aggarwal2019black}
Aniya Aggarwal, Pranay Lohia, Seema Nagar, Kuntal Dey, and Diptikalyan Saha.
\newblock Black box fairness testing of machine learning models.
\newblock In \emph{Proceedings of the 2019 27th ACM Joint Meeting on European
  Software Engineering Conference and Symposium on the Foundations of Software
  Engineering}, pages 625--635, 2019.

\bibitem[Agarwal et~al.(2022)Agarwal, Agarwal, and
  Agarwal]{agarwal2022fairness}
Avinash Agarwal, Harsh Agarwal, and Nihaarika Agarwal.
\newblock Fairness score and process standardization: framework for fairness
  certification in artificial intelligence systems.
\newblock \emph{AI and Ethics}, pages 1--13, 2022.

\bibitem[Kazim et~al.(2021)Kazim, Koshiyama, Hilliard, and
  Polle]{kazim2021systematizing}
Emre Kazim, Adriano~Soares Koshiyama, Airlie Hilliard, and Roseline Polle.
\newblock Systematizing audit in algorithmic recruitment.
\newblock \emph{Journal of Intelligence}, 9\penalty0 (3):\penalty0 46, 2021.

\bibitem[Landers and Behrend(2022)]{landers2022auditing}
Richard~N Landers and Tara~S Behrend.
\newblock Auditing the ai auditors: A framework for evaluating fairness and
  bias in high stakes ai predictive models.
\newblock \emph{American Psychologist}, 2022.

\bibitem[Dua and Graff(2017)]{dua2019}
Dheeru Dua and Casey Graff.
\newblock {UCI} machine learning repository, 2017.
\newblock URL \url{http://archive.ics.uci.edu/ml}.

\end{thebibliography}

\end{document}